# Image Classification by Feature Dimension Reduction and Graph based Ranking

YAO Nan[1], QIAN Feng[1] and SUN Zuolei[2]

[1] Department of Instrumentation Engineering, Shanghai Jiao Tong University, Shanghai 200240, China.

[2] College of Information Engineering, Shanghai Maritime University, Shanghai 201306, China

**Abstract.** Dimensionality reduction (DR) of image features plays an important role in image retrieval and classification tasks. Recently, two types of methods have been proposed to improve the both the accuracy and efficiency for the dimensionality reduction problem. One uses Non-negative matrix factorization (NMF) to describe the image distribution on the space of base matrix. Another one for dimension reduction trains a subspace projection matrix to project original data space into some low-dimensional subspaces which have deep architecture, so that the low-dimensional codes would be learned. At the same time, the graph based similarity learning algorithm which tries to exploit contextual information for improving the effectiveness of image rankings is also proposed for image class and retrieval problem. In this paper, after above two methods mentioned are utilized to reduce the high-dimensional features of images respectively, we learn the graph based similarity for the image classification problem. This paper compares the proposed approach with other approaches on an image database.

## Introduction

In modern days, on the Internet digital images are growing significantly fast, as the digital cameras become more and more widespread and affordable. In this situation, the automatically image classification, has become quite a significant topic in computer vision recently. It aims at classifying images according to their semantic labels automatically using computers [1]. In most of the image classification systems, the low-level visual features are usually extracted first for image classification and retrieval. Unfortunately, most effective visual features are at high-dimensional data space [2]. This problem is summarized as the so-called "curse of dimensionality." It has introduced several technical challenges including computational complexity, sparsity and redundancy. Many studies have been conducted for dimensionality reduction [3]. The high dimensionality is usually transformed into low dimensionality by minimizing the loss of the information which is contained by the high-dimensional data. The Dimensionality reduction methods can be generally classified into two types:

*Part based dimensionality reduction*: such as sparse coding [4] and nonnegative matrix factorization (NMF) [5]. It try to represent the data by the linear combination of a small number of basic elements, and the combination coefficients will be used as low dimensional data. Many studies have been done in this direction. For example, Terashima et al. [6] applied sparse coding for harmonic vocalization in monkey auditory cortex. Wang et al. [7, 8] has proposed the adaptive graph and multiple graph regularized NMF for data presentation work.

*Subspace Learning based dimensionality reduction*: such as Principal component analysis (PCA) [9] and Linear discriminant analysis (LDA) [10]. Lots of researches have been done in this field. For example, Wang et al. [11] learn the local subspace metrics which are optimized for local subset of the whole dataset, and Tzimiropoulos [12] introduced subspace learning from image for gradient orientations of image appearance for object recognition.

At the same time, graph base ranking has also been proposed for image classification and retrieval. It learns the similarity or distance metric [13] by help of the nearest neighbor graph. Some works have done to improve the graph based ranking methods. For example, Wang et al. [14] proposed the multiple graph regularized ranking for protein domain indexing problem. Matsuda et al. [15] applied the manifold ranking to multiple-food recognition problem by considering co-occurrence. Wang et al. have proposed the shortest path propagation [16] and the coherent dissimilarity-hierarchical context learning algorithm [17] to further improve the graph based ranking method.

In this paper, we try to propose a novel image classification method by combining the DR methods and the graph based ranking methods together. To this end, we need to answer the following questions:
1. which DR method should be used?
2. which graph ranking method should be used?
3. what's the optimal combination?

We investigate varies methods on a collected image database and use the one with the best classification result to design our system.

The rest parts of this paper are organized as follows: In method section, we introduce the proposed method and the database used for evaluation. In result section, the experiment results are reported. In the conclusion section, we conclude this paper with some insights.

**Proposed Approach**

In this section we will introduce the proposed approach and the database used to evaluate it.

**Image database.** We collect an image database from Internet with 1,000 images. The images are classified into 20 classes, and each class contains 50 images. Each image is labeled manually. Some images are shown in Fig. 1.

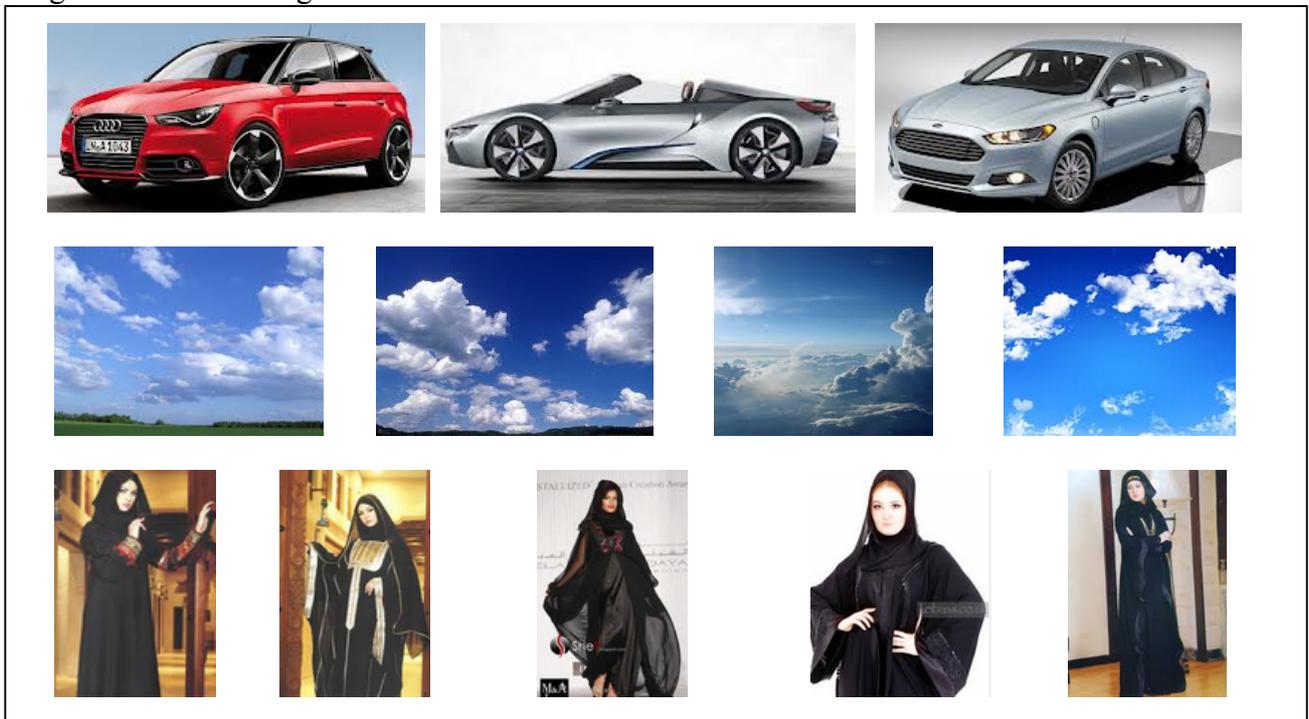

Fig. 1. Example images of the database.

**Proposed image classification approach.** The image classification system is shown in Fig. 2. The color, texture and shape features of each image are firstly extracted and combined as a high-dimensional visual feature vector, and then two dimensionality reduction methods, e.g. NMF and PCA are applied to the visual feature vector and combined to obtain the new image representation. Finally, the graph based similarity will be learned based on the new representation for the nearest neighbor classification.

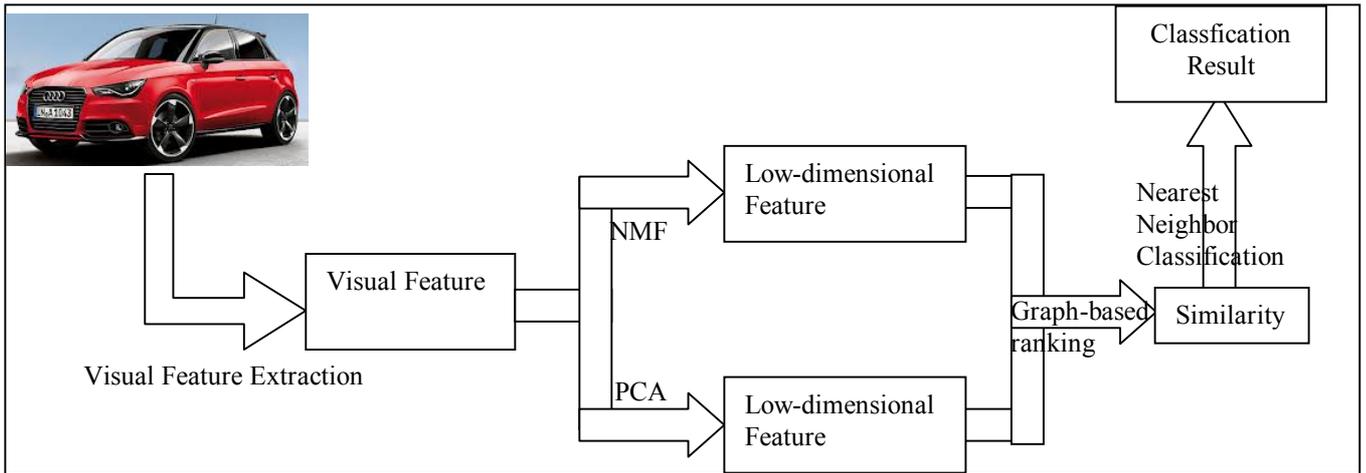
Fig. 2. Pipeline of the proposed approach.

**Experiments**

To test the proposed approach in the database, we applied the 10-fold cross-validation. The database is split into 10 folds, and each fold will be used as test set while the other ones as training set. The test will be repeated for 10 times. The average classification rate is used as the performance measure. The test results are given in Table 1. As we can see that by combining NMF, PCA for dimensionality reduction and graph ranking for similarity learning, the performance could be improved significantly.

Table 1. Average classification rates of the varies methods on the database.

| Methods | NMF | PCA | NMF+PCA | Graph ranking | NMF+PCA+Graph ranking |
|---|---|---|---|---|---|
| Average classification rate | 86.5% | 82.1% | 88.4% | 87.3% | 93.8% |

**Conclusion**

In this paper, we proposed a novel image classification approach by combing Dr and graph ranking methods. The results show that the proposed can improve the classification performance significantly.